\pdfoutput=1

\documentclass[11pt]{article}

\usepackage[]{acl}

\usepackage{times}
\usepackage{latexsym}

\usepackage[T1]{fontenc}

\usepackage[utf8]{inputenc}

\usepackage{microtype}

\usepackage{inconsolata}

\usepackage{graphicx}
\usepackage{amsmath}
\usepackage{float}
\usepackage{stfloats}
\usepackage{amsfonts}
\usepackage{colortbl}
\usepackage{booktabs,arydshln}
\usepackage{multirow}
\usepackage{adjustbox}
\usepackage{soul,color}

\newcommand{\cdashlinelr}[1]{%
  \noalign{\vskip\aboverulesep
           \global\let\@dashdrawstore\adl@draw
           \global\let\adl@draw\adl@drawiv}
  \cdashline{#1}
  \noalign{\global\let\adl@draw\@dashdrawstore
           \vskip\belowrulesep}}

\title{Accurate and Well-Calibrated ICD Code Assignment Through Attention Over Diverse Label Embeddings}

\author{Gonçalo Gomes \quad Isabel Coutinho \quad Bruno Martins \\  \textit{Instituto Superior Técnico and INESC-ID, University of Lisbon} \\ {\tt \{goncaloecgomes,isabel.coutinho,bruno.g.martins\}@tecnico.ulisboa.pt}}

\begin{document}
\maketitle

\begin{abstract}
Although the International Classification of Diseases (ICD) has been adopted worldwide, manually assigning ICD codes to clinical text is time-consuming, error-prone, and expensive, motivating the development of automated approaches. This paper describes a novel approach for automated ICD coding, combining several ideas from previous related work. We specifically employ a strong Transformer-based model as a text encoder and, to handle lengthy clinical narratives, we explored either (a) adapting the base encoder model into a Longformer, or (b) dividing the text into chunks and processing each chunk independently. The representations produced by the encoder are combined with a label embedding mechanism that explores diverse ICD code synonyms. Experiments with different splits of the MIMIC-III dataset show that the proposed approach outperforms the current state-of-the-art models in ICD coding, with the label embeddings significantly contributing to the good performance. Our approach also leads to properly calibrated classification results, which can effectively inform downstream tasks such as quantification.
\end{abstract}

\section{Introduction}

The International Classification of Diseases (ICD\footnote{\url{https://www.who.int/standards/classifications/classification-of-diseases}}) coding system, proposed by the World Health Organization, stands as a universal standard for precise documentation in the medical domain~\citep{O'Malley05}. Still, the manual assignment of ICD codes to clinical text is a time-consuming, labor intensive, and error-prone task, which has led to the exploration of automated methods, e.g. using deep learning algorithms for text classification.

Despite many previous efforts, automatic ICD coding is still challenging. Clinical notes consist of long text narratives that use a specialized medical vocabulary and are associated with a high dimensional, sparse, and imbalanced label space. 

In addition to accurately classifying individual clinical notes, estimating the prevalence of ICD codes within a dataset is also important for many practical applications. This corresponds to a text quantification problem~\citep{schumacher2021comparative,Moreo22}, for which access to properly calibrated text classification models can be helpful.

This paper describes a novel approach to ICD coding, combining several ideas from previous work. In particular, we explored two different text encoding strategies using a strong Transformer-based model~\citep{Yang22}, explicitly dealing with the lengthy nature of documents like hospital discharge summaries. The resulting representations are combined with a label embedding mechanism inspired by the proposal from \citet{Yuan22}, which explores diverse ICD code synonyms. Additionally, taking inspiration from the MLP-based quantification approach from \citet{coutinho2023exploring}, we present a training setup in which multi-label classification and text quantification are jointly addressed. This additional step aims to improve model calibration while also informing downstream tasks such as text quantification.

Following previous studies, the proposed model was evaluated on the publicly available MIMIC-III dataset \citep{Johnson16}, specifically analyzing results on two subsets of hospital discharge summaries, namely MIMIC-III-50 \citep{Mullenbach18} and MIMIC-III-clean \citep{Edin23}. Our approach surpasses standard baselines and previous state-of-the-art models for ICD coding across all evaluated metrics, while simultaneously providing interesting results regarding model calibration and ICD code quantification. The source code supporting our experiments is available in a GitHub repository\footnote{\url{https://github.com/gecgomes/ICD_Coding_MSAM}}.

The remaining parts of this paper are organized as follows: Section 2 reviews the related literature, while Section 3 introduces our novel approach for ICD coding and quantification. Section 4 presents the experimental results, establishing a direct comparison with previous studies. Finally, Section 5 summarizes our contributions and discusses future research directions. The paper ends with a discussion on limitations and ethical considerations.

\section{Related Work}

Several previous studies have addressed the problem of automatic ICD coding. For instance, \citet{Mullenbach18} introduced the Convolutional Attention for Multi-Label classification (CAML) approach, which is still commonly considered as a baseline. CAML employs a label-wise attention mechanism that enables the model to learn distinct document representations for each label, selecting relevant parts of the document for each ICD code. The authors conducted experiments on MIMIC datasets \citep{Lee11,Johnson16}, and the data splits developed for this work were made publicly available. This study is considered an essential milestone for reproducibility.

Aiming to address CAML's limitations in capturing variable-sized text patterns, \citet{Xie19} improved the convolutional attention model by introducing a densely connected CNN with multi-scale feature attention (MSATT-KG), which produces variable $n$-gram features and adaptively selects informative features based on neighborhood context. This method also incorporates a graph CNN to capture hierarchical relationships among medical codes. In turn, \citet{Li20} proposed MultiResCNN, i.e. a tailored CNN architecture combining multi-filter convolutions and residual convolutions, capturing patterns of different lengths and achieving superior performance over CAML.

\citet{Vu21} introduced LAAT, i.e., a model that combines an RNN-based encoder with a new label attention mechanism for ICD coding. LAAT aimed to handle the variability in text segment lengths and the interdependence among different segments related to ICD codes. Additionally, the authors introduced a hierarchical joint learning mechanism to address the class imbalance issue.

\citet{Yuan22} put forth the Multiple Synonyms Matching Network (MSMN) as an alternative approach to ICD coding. Rather than relying on the ICD code hierarchy, the authors leveraged synonyms to enhance code representation learning and improve coding performance. 

Nowadays, Transformer-based Language Models (LMs) are becoming the fundamental technology for medical AI systems to process clinical narratives. For instance, \citet{Yang22} developed a large clinical LM named GatorTron, training a Transformer encoder model on text narratives from de-identified clinical notes from University of Florida Health, PubMed articles, and Wikipedia. The authors also examined how increasing the number of parameters enhances the performance on various Natural Language Processing (NLP) tasks, including named entity recognition, medical relation extraction, semantic textual similarity, natural language inference, and medical question answering. Their results showed that GatorTron outperformed previous Transformer models across various NLP tasks within the biomedical and clinical domains.

\citet{Dai22} compared Transformer models for long document classification, focusing on mitigating the computational overheads associated with encoding large texts. In turn, \citet{Huang22} investigated limitations associated with using pre-trained Transformer-based LMs, identifying challenges regarding large label spaces, long input lengths, and domain disparities. The authors proposed PLM-ICD, i.e., a framework that effectively handles these challenges and achieves superior results on the MIMIC-III dataset, surpassing previously existing methods.

In a recent study, \citet{Edin23} argued that the proper assessment of model performance on ICD coding had often struggled with weak configurations, poorly designed train-test splits, and inadequate evaluation procedures. The authors pinpointed significant issues with the MIMIC-III splits released by \citet{Mullenbach18}. They proposed a new dataset split using stratified sampling to ensure a complete representation of all classes, referred to as MIMIC-III-clean.

Regarding text quantification, various algorithms have been proposed in recent years~\citep{schumacher2021comparative}. Still, few previous studies have specifically considered multi-label settings~\citep{Moreo22}. \citet{coutinho2023exploring} explored the use of a Multi-Layer Perceptron (MLP) model for ICD code quantification, taking inspiration from under-complete denoising auto-encoders. The MLP was trained to refine estimates provided by the Probabilistic Classify and Count (PCC) method, considering label correlations. Experiments with different MIMIC-III dataset splits showed that the proposed method outperforms baseline approaches such as Classify and Count (CC) and PCC. 

\section{Proposed Approach}

This work presents a novel approach for ICD coding, aiming at strong classification performance together with well-calibrated outputs, which can inform downstream tasks such as text quantification.

\subsection{Clinical Text Modeling}
\label{subsection: text_modeling}
Within the proposed approach, we compared two different text encoding strategies to handle long clinical documents, using GatorTron-base as the foundational Language Model (LM) for both strategies. GatorTron-base is a Megatron BERT model pre-trained on the healthcare domain, previously described by \citet{Yang22}. This model is publicly available in the NVIDIA NGC\footnote{\url{https://catalog.ngc.nvidia.com/}} Catalog and also through the HuggingFace\footnote{\url{https://huggingface.co/UFNLP/gatortron-base}} library.

As a first strategy, we considered a Longformer Encoding (LE) approach, where the standard self-attention mechanism of GatorTron-base is replaced by the Longformer sparse attention mechanism, which scales linearly with sequence length~\cite{beltagy2020longformer}. This makes it possible to process documents of thousands of tokens. To adapt the GatorTron-base into a Longformer, we resized the positional embeddings to match the new maximum length allowed for the input sequences (in our case, equal to $8,192$ tokens). The LE uses a local windowed self-attention over chunks of $512$ tokens, and global self-attention in the {\tt [CLS]} token.

As a second strategy, we considered a Chunk Encoding (CE) approach, dividing the documents into $C$ chunks and processing them individually with GatorTron-base. By dividing inputs into chunks, we can effectively leverage the capabilities of a standard Transformer encoder, limited to a maximum of $T$ tokens (in our case, $T = 512$), to analyze long clinical documents. While both LE and CE are simple to implement, CE can perhaps better avoid difficulties in having LMs making robust use of information within long inputs~\cite{liu2023lost}.

\begin{figure}[t!]
  \centering
 \includegraphics[width=0.9\columnwidth]{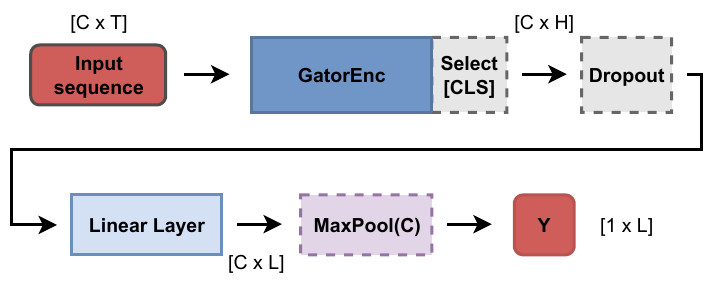}
  \vspace{-0.25cm}
  \caption{A simple classification architecture that considers the Chunk Encoding (CE) approach.}
\label{fig:architecture}
\end{figure}

\begin{figure}[t!]
  \centering
  \vspace{-0.3cm}
  \includegraphics[width=0.9\columnwidth]{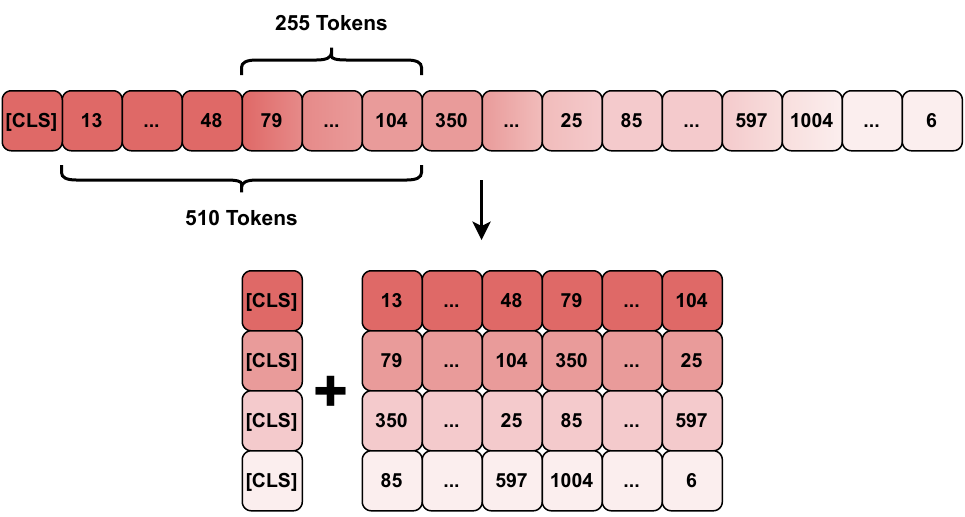}
  \vspace{-0.25cm}
  \caption{Smooth document segmentation with 255 token overlaps. Each chunk includes, at the end, the sentence separation token {\tt [SEP]} characteristic of BERT-type models, completing $512$ tokens per chunk.}
\label{fig:smoothslice}
\end{figure}

CE relies on the assumption that if an ICD code can be identified in a single segment (i.e., a chunk) of the input document, then the code should naturally also be assigned when classifying the document as a whole (i.e., a single mention should be enough to justify the coding decision). We thus use a max-pooling operation to consolidate the detection of different ICD codes in each chunk, as illustrated in Figure~\ref{fig:architecture}, where $C$ refers to the number of chunks, $T$ corresponds to the number of tokens within each chunk, $H$ corresponds to the dimensionality of the vectors representing each token, and $L$ denotes the number of ICD classes. We also adopted a smooth partitioning scheme that considers overlaps between chunks to mitigate the loss of information from abruptly breaking interconnected pieces of text, as shown in Figure~\ref{fig:smoothslice}. 

In the remaining parts of the paper, we will collectively refer to GatorTron and its Longformer version as the GatorTron encoder (GatorEnc). Note that when using the LE strategy, the number of chunks $C$ can be seen as being equal to one, and we do not require the max-pooling operation to consolidate chunk results. With this, we can interpret the following images and expressions in a way that generalizes to both the LE and CE approaches.

\subsection{Multi-Synonyms Attention}
Inspired by \citet{Yuan22}, we enhanced our classification model by integrating a multi-synonyms attention mechanism. The primary objective was to explore the intricate relationships between specific mentions of ICD codes, within chunks of the hospital discharge summaries, and the textual descriptions of ICD codes. This integration aimed to leverage synonyms to improve code representation learning (i.e., label embeddings), aiding in code classification.

First, we extended the ICD code descriptions with synonyms obtained from a large medical knowledge base, specifically the UMLS\footnote{\url{https://www.nlm.nih.gov/research/umls/}} meta-thesaurus. By aligning ICD codes with UMLS Concept Unique Identifiers (CUIs), we selected corresponding synonyms for English terms sharing the same CUIs. Additionally, we considered synonym variants by removing special characters, allowing only hyphens and brackets, and removing the coordinating conjunctions {\it or} and {\it and}.

To improve diversity, we also gathered additional synonyms from Wikidata and Wikipedia. However, even with this addition, we observed that the lists of synonyms associated with each ICD code were often repetitive, posing a risk of introducing undesired bias in classification. We selected a maximum of $M$ synonyms by first representing them as vectors through GatorEnc (i.e., taking the {\tt [CLS]} token representation for each synonym). Then, a set of $M$ diverse vectors were selected for each ICD code through the application of the Gurobi optimizer\footnote{\url{https://www.gurobi.com}} as a way to address the Maximum Diversity Problem\footnote{\url{https://grafo.etsii.urjc.es/optsicom/mdp.html}} (MDP), introduced by~\citet{glover1977selecting} and which can be formulated as follows:
\begin{equation}
\vspace{-0.1cm}
\scalebox{0.925}{$
    \text{maximize} \sum_{i = 1}^{N-1} \sum_{j = i + 1}^{N} d_{ij}x_ix_j, \text{}
    $}
\end{equation}
\begin{equation}
\scalebox{0.925}{$
     \text{subject to} \sum_{i = 1}^{N} x_i = M, \text{}
     $}
\end{equation}
\begin{equation}
\scalebox{0.925}{$
     x_i = \{0,1\}, \quad 1\leq i \leq N.
     $}
\vspace{-0.1cm}
\end{equation}
In the previous equations, $d_{ij}$ is a distance metric between synonym representations $i$ and $j$ (i.e., the cosine distance between the vectors), and $x_i$ takes the value one if element $i$ is selected and 0 otherwise. By solving the MDP, we select a small subset of $M$ out of $N$ synonyms, that effectively represent the broader embedding space for each ICD code. 

We can denote by $Q_l$ a matrix where rows correspond to the representations for the $M$ synonyms associated to an ICD code $l$, with each code synonym $j_l$ composed of tokens $\{s_i^{j_l}\}^{S_{j_l}}_{i=1}$:
\begin{equation}
\vspace{-0.1cm}
    Q_l = \{\text{GatorEnc(}s_1^{j_l},...,s_{S_{j_l}}^{j_l}\text{){\tt [CLS]}}\}^M_{j_l=1}.
\vspace{-0.1cm}    
\end{equation}
Note that the synonym representations are not updated during model training. The token representations of hidden size $H$, within each chunk of text $c$, are similarly produced with GatorEnc, as follows:
\begin{equation}
    K^c = \text{GatorEnc(}x_1^c,...,x_{T}^c\text{)}.
\end{equation}
Thus, $K$ corresponds to the aggregate token representations of all chunks. To integrate the text representations from each chunk with the multiple synonym representations, we use an approach inspired by the multi-synonyms attention method proposed by \citet{Yuan22}, which in turn draws inspiration from the multi-head attention mechanism of the Transformer architecture~\citep{Vaswani17}. We specifically split each $K^c$ into $Z$ heads, equaling this value to the maximum number of synonyms per code, i.e., $Z=M$: 
\begin{equation}
    K^c = K^c_1, ... , K^c_Z.
\end{equation}
The code synonyms $\{Q_l\}^{L}_{l=1}$ are used to query each $K^c$ and, by calculating attention scores $\alpha_l$ over $K^c$, we identify the parts from the chunk's text that are more related to code synonym $l$: 
\begin{equation}
    \alpha_l = \{\text{Softmax(}W_QQ_l \thinspace.\thinspace \text{Tanh(}  W_KK^c\text{)}\text{)}\}^C_{c=1}.
\end{equation}
We use an average pooling operation over $\text{Tanh(}K\text{)}\alpha_l$ to create code-wise text representations $R$, averaging the contributions from synonyms:
\begin{equation}
    R = \{\text{AvgPool(}\text{Tanh(}K\text{)}\alpha_l\text{)}\}^L_{l=1}.
\end{equation}
To assess whether the text of a chunk $c$ contains code $l$, we evaluate the similarity between the code-wise text representation $R_c$ and the code's embeddings $V$. We aggregate the synonym representations {$\{ Q_l \}_{l=1}^L$} to form code representations $V$ through average pooling, resulting in a matrix with each row depicting a global code representation:
\begin{equation}
V = \{ \text{AvgPool(}Q_l^1,Q_l^2,...,Q_l^M\text{)} \}_{l=1}^L.
\end{equation}
To measure the similarity for classification, we apply a bi-affine transformation. Finally, after carefully attending to the ICD codes in each chunk, using synonyms to enhance the classification, we employ max-pooling to consolidate the results:
\begin{equation}
\scalebox{0.8}{$
Y = \sigma\text{(} \text{MaxPool(} \text{Diag(}R_1'WV\text{)},...,\text{Diag(}R_C'WV\text{)} \text{)} \text{)}.
$}
\end{equation}
In the previous equation, $R_c'$ is the transpose of $R_c$. Unlike previous approaches that perform classification using learned code-dependent parameters, which can be challenging to define for rare codes, our bi-affine function uses the parameters $WV$, where $W$ is learned. This simplifies the learning process, at the same time making it more effective.  

\begin{figure}
  \centering
  \includegraphics[width=0.9\columnwidth]{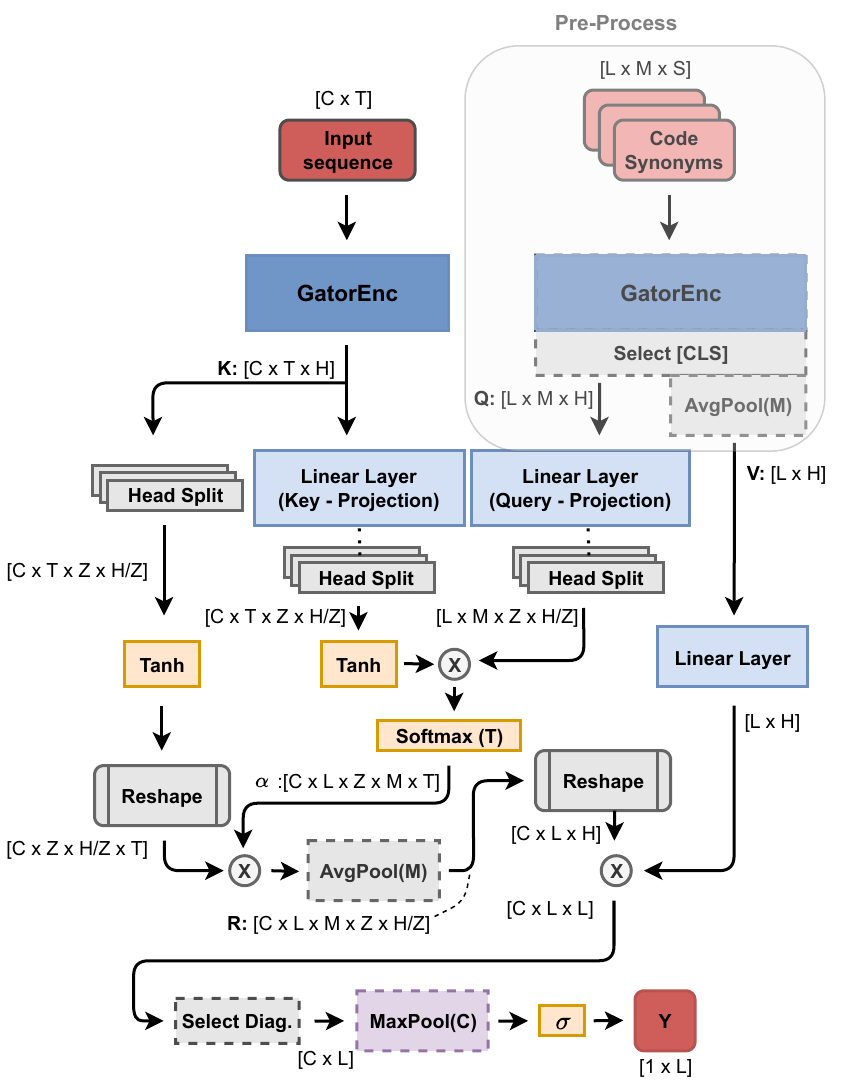}
  \vspace{-0.25cm}

  \caption{The classification architecture that combines the CE with a multi-synonyms attention mechanism.}
  \label{fig:seg_megatron_att}
\end{figure}

Figure \ref{fig:seg_megatron_att} illustrates the combination of the chunk-based encoding strategy, described in the previous section, with the classification method that considers the multi-synonyms attention mechanism.

For model training, noting that we are in the presence of a multi-label classification task, we adopted the widely-used Binary Cross-Entropy (BCE) loss, which treats each class independently and can be formally described as follows:
\begin{equation}
\label{equation: BCE}
\scalebox{0.9}{$
    \mathcal{L}_{C} = \sum_{l = 1}^L -y_l\text{log(}\hat{y}_l\text{)} - (1-y_l)\text{log}(1-\hat{y}_l\text{)}.
    $}
\end{equation}
The variable $y_l \in \{0,1\}$ represents the ground-truth for a code $l$, while $\hat{y}_l$ represents the probability of that code being present, as given by the classifier, and $L$ is the number of different ICD codes.

\subsection{Joint Classification and Quantification}
\label{subsection: CLQ}

Following previous work by \citet{coutinho2023exploring}, we considered an approach inspired by under-complete denoising auto-encoders to quantify the prevalence of ICD codes within a set of documents, accounting with label associations. We integrated this quantification module, implemented as a three-layer MLP, together with the classifier, performing end-to-end training of the resulting model. We hypothesize that the classification and the quantification objectives can naturally complement each other, and that combining them can contribute to a better calibration of the model.

Notice that classification operates at the level of individual instances, while quantification operates over groups of instances. To integrate both objectives within end-to-end training, we follow the steps described next:
\begin{enumerate}
\label{enumeration: classy&quant}
    \item \textbf{Shuffling and setting a limit:} We shuffle the dataset at the start of each training epoch. We also establish a limit that simulates the maximum number of instances to be considered for quantification, selected randomly between one and the total number of training instances.

    \vspace{-0.15cm}    
    \item \textbf{Iterative data collection:} We process the instances individually as we progress through the training set. We collect the classification results for each instance that is processed, until we hit the previously defined maximum limit. This creates a new group of instances for each instance that is processed, consisting of all the instances that we have processed thus far, plus the latest one. The processing of each instance is made as follows: 

    \vspace{-0.15cm}    
    \begin{enumerate}    
        \item \textbf{Computation of classification loss:} When processing each new instance, we apply our classification model and calculate the classification loss associated with that instance.
        \item \textbf{Computation of quantification loss:} We add the classification output to the previous classification outputs for instances within the group. This allows us to compute a Probabilistic Classify and Count (PCC) vector, denoting the estimated relative frequency of each class label within the group of instances. We then process this vector using the aforementioned MLP, which refines the PCC estimates. We finally calculate the quantification loss with the refined estimates.
        \item \textbf{Aggregation of results:} The loss values computed in the previous steps are aggregated into a total loss, which is used to update the model parameters for each batch of processed instances.
    \end{enumerate}

    \vspace{-0.15cm}
    \item \textbf{Repeat and reset:} We follow the iterative process (i.e., steps (a) to (c)) until we reach the maximum number of instances designated for the quantification set. Once this limit is reached, we reset the quantification group and establish a new maximum limit for quantifying instances. We continue with model training until a stopping criterion is met.
\end{enumerate}

The proposed approach performs joint training with classification and quantification, accumulating the PCC quantification estimates over batches of instances to produce refined quantification results for groups of different sizes. At inference time, we can take the MLP that was trained jointly, and use it separately to perform quantification (i.e., a classifier can process each instance in a group in order to produce class probabilities, which can be aggregated into a PCC quantification estimate to be then refined with the MLP). 

Our combined loss function can be formally described by the following equation, where $\lambda$ is a hyper-parameter controlling the relative influence of the quantification loss:
\begin{equation}
\label{equation: classi_quant}
    \vspace{-0.1cm}
    \mathcal{L} = \mathcal{L}_C + \lambda \mathcal{L}_Q.
    \vspace{-0.1cm}    
\end{equation}
The classification loss ($\mathcal{L}_C$) is the BCE, formally described in Equation \ref{equation: BCE}, and the quantification loss ($\mathcal{L}_Q$) can correspond to either the common Mean Squared Error (MSE) loss ($\mathcal{L}_Q^{MSE}$), or the Huber loss ($\mathcal{L}_Q^{Huber}$), respectively given by:
\begin{equation}
\scalebox{0.9}{$
    \mathcal{L}_{Q}^{MSE}(\hat{p}_\epsilon,p_\epsilon) = \sum_{l = 1}^L (\Delta_l)^2,
    $}
\end{equation}
\begin{equation}
\scalebox{0.9}{$
    \mathcal{L}_{Q}^{Huber}(\hat{p}_\epsilon,p_\epsilon) = \begin{cases}
      \frac{1}{2}\sum_{l = 1}^L (\Delta_l)^2 &\text{if $\Delta_l < \delta$,}\\
      \delta \cdot (\Delta_l - \frac{1}{2}\delta)  &\text{otherwise.}
    \end{cases}
    $}
\end{equation}
In the previous expressions, $\Delta_l = |\hat{p}_\epsilon(l) - p_\epsilon(l)|$, where $p_\epsilon$ refers to the ground-truth quantification result (i.e., the relative class frequency within the set of instances) for each of the $L$ class labels, $\hat{p}_\epsilon$ refers to the quantification estimates, and $\delta$ is a tuning parameter that determines the point at which the Huber loss transitions from a quadratic to a linear penalty. The Huber loss has a smooth optimization landscape, like the MSE, but it is less sensitive to outliers, thus perhaps leading to more stable and accurate results.

\section{Experimental Evaluation}
This section presents the experimental evaluation of the proposed method, establishing a comparison with previously reported results.

\subsection{Datasets}
Experiments were conducted using the publicly available MIMIC-III data \cite{Johnson16}, which we accessed through PhysioNet\footnote{\url{https://physionet.org/content/mimiciii/}} after completing the ethical training program from the associated collaborative institutional training initiative. We specifically used the same dataset splits considered in previous studies, namely MIMIC-III-50 \cite{Mullenbach18}, which only comprises the top-50 most frequent ICD-9 codes in the dataset, and also MIMIC-III-clean \cite{Edin23}, which corresponds to a cleaned version of the dataset that contains 3,681 unique ICD-9 codes. We present a statistical characterization of the dataset splits in Appendix \ref{appendix: dataset}. 

The quantification experiments also used MIMIC-III-50 and MIMIC-III-clean, following the general methodology from \citet{coutinho2023exploring}. Specifically, to assess result quality in each case, we sampled documents from the corresponding validation set to form $5,000$ quantification groups of different sizes, with the size parameter varying between one and the number of documents in the set. A separate set of $1,000$ groups was also created by sampling documents from the test split. These sets were used for model training and testing in the quantification experiments.  

\subsection{Evaluation Metrics}
We assessed the proposed approach across various metrics considered in previous work, to ensure a fair comparison with prior research.

Regarding the classification task, we used micro and macro-averaged F1 scores, Area Under the Curve (AUC) scores, and precision at cutoff $n$. For the experiments over the MIMIC-III-50 dataset, we defined $n = 5$. For the experiments conducted on MIMIC-III-clean, we considered $n = 8$ and $n = 15$, roughly aligning with the average number of codes in each split. To measure our classifier's calibration quality, we used the Mean Expected Calibration Error (MECE) with $20$ bins.

To evaluate the quantification task, we used the Mean Absolute Error (MAE) and the Mean Relative Absolute Error (MRAE) \citep{coutinho2023exploring}. The MRAE uses an additive smoothing penalty to avoid divisions by zero, which slightly impacts results according to the size of the groups.

\subsection{Implementation Details}
Table \ref{table: hyper-parameters} presents the training hyper-parameters considered in our experiments.

\begin{table}[!t]
\renewcommand{\arraystretch}{1.25}
\begin{adjustbox}{width=\columnwidth,center}
\begin{tabular}{lcc}
\hline
\rowcolor[HTML]{FFFFFF} 
\textbf{Parameters}                          & \textbf{MIMIC-III-50} & \textbf{MIMIC-III-clean} \\ \hline
\rowcolor[HTML]{F3F3F3} 
Maximum token input length for CE   & $7,142$                 & $6,122$                    \\
Maximum token input length for LE   & $8,192$                 & $8,192$                    \\
\rowcolor[HTML]{F3F3F3} 
Token overlapping window for CE     & $255$                   & $255$                      \\
GatorEnc hidden size ($H$)            & $1,024$                 & $1,024$                    \\
\rowcolor[HTML]{F3F3F3} 
Synonyms per ICD code ($M$)           & $4$                     & $4$                        \\
Number of heads ($Z$)                 & $4$                     & $4$                        \\
\rowcolor[HTML]{F3F3F3} 
Maximum number of epochs            & $300$                   & $300$                      \\
Early stopping patience             & $5$                     & $5$                        \\
\rowcolor[HTML]{F3F3F3} 
Effective batch size                & $16$                    & $16$                       \\
Adam $e$                              & $1$e-$08$                 & $1$e-$08$                    \\
\rowcolor[HTML]{F3F3F3} 
Starting learning rate              & $2$e-$05$/$2$e-$07$           & $2$e-$05$/$2$e-$07$              \\
Ending learning rate                & $0$                     & $0$                        \\
\rowcolor[HTML]{F3F3F3} 
MLP hidden size                     & $32$                    & $3,072$                    \\
Quantification coefficient ($\lambda$) & $100$                   & $100$      \\ \hline               
\end{tabular}
\end{adjustbox}
\caption{Hyper-parameters used for model training in the MIMIC-III-50 and MIMIC-III-clean settings. The \textit{maximum number of epochs} values are related to both the classification and quantification modules.}
\label{table: hyper-parameters}
\end{table}

Models using the Chunk Encoder (CE) have a maximum allowed number of input tokens limited only by hardware constraints. During training, we had to limit this input length according to the available GPU memory, considering a single NVIDIA A100 with 80Gb. However, at inference time, we could increase this limit up to $20,000$ tokens. In turn, models using the Longformer Encoder (LE) have the same maximum input length for training and testing. Once again, we set this value based on the available GPU memory.

We trained our classifiers in two stages, using a linear scheduler for the learning rate. The first stage uses a learning rate starting at $2e$-$5$ and proceeds until we reach an early stopping criteria based on the micro-averaged F1 score over the development set. We then perform a second training stage, with a learning rate starting at $2e$-$7$ and an early stopping criteria based on the Mean Expected Calibration Error (MECE). The quantification model (i.e., the MLP) was first trained individually following the guidelines of \citet{coutinho2023exploring}, using a learning rate that starts at $2e$-$5$ and proceeding until we reach an early stopping criterion, based on the MSE loss. The model that integrates the quantification objective was initialized with pre-trained classification and quantification components obtained after a first stage of training. Thus, these components should already perform each task with reasonable competence, prior to their combination.

\subsection{Experiments and Results}

We comprehensively evaluated the proposed approach with different metrics, comparing it against previous methods and ablated model versions. 

Our Baseline Model (BM) uses GatorTron-base to process the first $512$ tokens of each document, without the Longformer Encoder (LE) or Chunk Encoder (CE) strategies, and without the Multiple-Synonyms Attention Mechanism (MSAM). We also assessed the combination of the two encoding strategies with the label embeddings (i.e., models referred to as \{CE, LE\}+MSAM), and also the joint training with classification and quantification objectives (i.e., \{CE, LE\}+MSAM+CLQ).

\subsubsection{Classification}

Tables \ref{table: top50results} and \ref{table: cleanresults} present classification results for different model variants, respectively for MIMIC-III-50 and MIMIC-III-clean.

Both encoding strategies (CE and LE) significantly outperform the Baseline Model (BM), with CE also outperforming LE in all metrics by a considerable margin. Notice that LE compresses the entire document into a single representation vector, while the CE strategy considers identifying ICD codes in smaller chunks, avoiding problems in encoding long inputs~\cite{liu2023lost}. 

The MSAM mechanism also notably enhances performance across all the metrics. Still, despite the significant performance gap between the LE and CE models, these differences diminish after incorporating this module. In MIMIC-III-clean, LE+MSAM has a slightly higher micro-averaged F1 score (+$0.6$) but a significantly lower macro-averaged F1 score (-$2.5$) when compared to CE+MSAM. This indicates that while both versions show similar overall results across the different ICD classes, CE performs better in classifying individual ICD classes with lower prevalence frequencies, which can be important in this domain. 

Considering the overall better results with the CE strategy, we decided to use only this encoding approach for further experiments, considering joint training with the quantification objective. The results in Tables \ref{table: top50results} and \ref{table: cleanresults} show that the joint training, either with the MSE or the Huber loss, does not significantly impact the classification accuracy.

\begin{table}[!t]
\renewcommand{\arraystretch}{1.2}
\begin{adjustbox}{width=\columnwidth,center}
\begin{tabular}{lcccccc}
\hline
\rowcolor[HTML]{FFFFFF} 
\cellcolor[HTML]{FFFFFF}                                 & \textbf{Stopping}                            & \multicolumn{2}{c}{\cellcolor[HTML]{FFFFFF}\textbf{AUC}} & \multicolumn{2}{c}{\cellcolor[HTML]{FFFFFF}\textbf{F1}} & \textbf{P@N} \\ \cline{3-7} 
\rowcolor[HTML]{FFFFFF} 
\multirow{-2}{*}{\cellcolor[HTML]{FFFFFF}\textbf{Model}} & \textbf{Epochs}                              & \textbf{Macro}              & \textbf{Micro}             & \textbf{Macro}             & \textbf{Micro}             & \textbf{P@5} \\ \hline
CAML* \citep{Mullenbach18}    &  $-$      & $87.5$            & $91.1$           & $51.0$           & $60.6$           & $61.1$            \\
MSATT-KG$^\dagger$ \citep{Xie19}  &  $-$      & $91.4$            & $93.6$           & $63.8$           & $68.4$           & $64.4$            \\
MultiResCNN* \citep{Li20}          &  $-$      & $89.7$            & $92.4$           & $61.1$           & $67.3$           & $64.4$            \\
LAAT* \citep{Vu21}                 &  $-$      & $90.5$            & $92.8$           & $59.2$         & $66.8$           & $64.0$            \\
PLM-ICD* \citep{Huang22}           &  $-$      & $91.7$            & $93.8$           & $65.4$           & $70.5$           & $65.7$            \\
MSMN$^\dagger$ \citep{Yuan22}      &  $-$      & $92.8$            & $94.7$           & $68.3$           & $72.5$           & $68.0$\\ 
KEPTLongformer$^\dagger$ \citep{YangKept22}  &  $-$      & $92.6$            & $94.8$           & $68.9$           & $72.9$           & $67.3$            \\ \hline
\rowcolor[HTML]{F3F3F3} 
BM                                 &  $11 (+0)$     & $83.8$            & $87.0$           & $49.1$           & $56.1$          & $55.8$            \\
LE                          &  $26 (+0)$ & $84.9$     & $87.9$    & $54.8$           & $61.5$           & $59.7$            \\
CE                                 &  $10 (+0)$     & $91.2$            & $93.4$           & $65.5$           & $70.0$          & $66.1$            \\ 
\rowcolor[HTML]{F3F3F3}
LE+MSAM                          &  $5 (+6)$ & \textbf{93.8}     & \textbf{95.4}    & $70.3$           & $73.9$           & \textbf{69.1}            \\
\rowcolor[HTML]{F3F3F3} 
CE+MSAM                          &  $4 (+10)$ & $93.7$     & \textbf{95.4}    & \textbf{70.4}           & $73.9$           & $68.8$            \\
CE+MSAM+$\text{CLQ}_\text{MSE}$                    &  $4 (+4)$ & $93.7$   & \textbf{95.4}           & \textbf{70.4}    & \textbf{74.0}    & $68.9 $  \\
 
CE+MSAM+$\text{CLQ}_\text{Huber}$    &  $4 (+1)$ & $93.7$       & \textbf{95.4}           & $70.3$          & $73.8$    & $68.9$        \\ \hline    
\end{tabular}
\end{adjustbox}
\caption{Results for the different classification methods on the MIMIC-III-50 test set. Results for methods marked with * were taken directly from \citet{Edin23}. Results for methods marked with $\dagger$ were taken directly from the corresponding paper. The values in bold represent the best-in-class performance in terms of the different evaluation metrics.}
\label{table: top50results}
\end{table}

\begin{table}[!t]
\renewcommand{\arraystretch}{1.25}
\begin{adjustbox}{width=\columnwidth,center}
\begin{tabular}{lccccccc}
\hline
\rowcolor[HTML]{FFFFFF} 
\cellcolor[HTML]{FFFFFF}                                 & \textbf{Stopping}    & \multicolumn{2}{c}{\cellcolor[HTML]{FFFFFF}\textbf{AUC}} & \multicolumn{2}{c}{\cellcolor[HTML]{FFFFFF}\textbf{F1}} & \multicolumn{2}{c}{\cellcolor[HTML]{FFFFFF}\textbf{P@N}} \\ \cline{3-8} 
\rowcolor[HTML]{FFFFFF} 
\multirow{-2}{*}{\cellcolor[HTML]{FFFFFF}\textbf{Model}} & \textbf{Epochs}      & \textbf{Macro}              & \textbf{Micro}             & \textbf{Macro}             & \textbf{Micro}             & \textbf{P@8}   & \cellcolor[HTML]{FFFFFF}\textbf{P@15}   \\ \hline
CAML* \citet{Mullenbach18}        &  $-$       & $91.4$          & $98.2$         & $20.4$         & $55.4$         & $67.7$        & $52.8$  \\
MultiResCNN* \citep{Li20}         &  $-$       & $93.1$          & $98.5$         & $22.9$         & $56.4$         & $68.5$        & $53.5$ \\
LAAT* \citep{Vu21}                &  $-$       & $94.0$          & $98.6$         & $22.6$         & $57.8$         & $70.1$        & $54.8$ \\
PLM-ICD* \citep{Huang22}          &  $-$       & $95.9$          & $98.9$         & $26.6$         & $59.6$         & $72.1$        & $56.5$ \\ \hline
\rowcolor[HTML]{F3F3F3} 
BM                                &  $43 (+0)$      & $89.9$          & $95.7$         & $11.0$         & $44.5$         & $59.5$        & $44.2$                 \\
LE                                &  $79 (+0)$      & $90.3$          & $95.7$         & $12.9$         & $48.6$         & $63.0$        & $46.9$ \\
CE                                &  $68 (+0)$      & $91.7$          & $96.1$         & $16.9$         & $52.1$         & $66.1$        & $50.6$ \\
\rowcolor[HTML]{F3F3F3}
LE+MSAM                   &  $6 (+6)$      & $96.3$          & $\textbf{99.0}$         & $28.0$         & \textbf{60.9}         & \textbf{74.2}        & \textbf{58.1} \\
\rowcolor[HTML]{F3F3F3} 
CE+MSAM                           &  $8 (+5)$  &   $96.3$   & $98.9$   & $30.5$  &  $60.3$  & $73.3$ & $57.5$ \\
CE+MSAM+$\text{CLQ}_\text{MSE}$                       &  $8 (+6)$  & \textbf{96.4}   & \textbf{99.0}  & \textbf{31.2}  & $60.5$  & $73.3$ & $57.4$\\
CE+MSAM+$\text{CLQ}_\text{Huber}$           &  $8 (+5)$  & $96.3$   & $98.9$  & $30.5$ & $60.4$  & $73.3$ & $57.4$\\ \hline
\end{tabular}
\end{adjustbox}
\caption{Results for the different classification methods on the MIMIC-III-clean test set. Results for methods marked with * were taken from \citet{Edin23}.}
\label{table: cleanresults}
\end{table}

To assess the impact of using a different number of synonyms in the label embeddings, and also the diversity-based strategy for selecting the synonyms, we considered the CE+MSAM model over the MIMIC-III-50 dataset. We varied $M$ between $2$, $4$, or $8$ synonyms, and considered either random or maximum-diversity (i.e., MDP) selection. The results are shown in Table~\ref{table:M-settings}, confirming that the maximum-diversity selection strategy positively impacts the results when using fewer synonyms. Consistently with the results from \citet{Yuan22}, our experiments also indicate that $M = 4$ produces the best results.

We also analyzed the proposed approach in terms of calibration performance. In Table \ref{table:MECE}, we explicitly examine the calibration error over different sets of ICD codes: Low percentile ({\it Low Pth}) corresponds to the average value of the calibration error calculated for the 10\% of ICD codes with the lowest frequency rates in the training set of the respective MIMIC-III split. In turn, the medium percentile ({\it Medium Pth}) represents the average value of the calibration error for the 10\% of ICD codes with medium frequency rates, falling within the 55\% to 65\% percentile range in the respective MIMIC-III split training set. Finally, the high percentile ({\it High Pth}) indicates the average value of the calibration error for the 10\% of medical codes with the highest frequency of occurrence in the training set of the respective MIMIC-III split. 

\begin{table}[t!]
\centering
\renewcommand{\arraystretch}{1.25}
\resizebox{0.90\columnwidth}{!}{%
\begin{tabular}{lccccc}
\hline
\rowcolor[HTML]{FFFFFF} 
\cellcolor[HTML]{FFFFFF}                                   & \multicolumn{2}{c}{\cellcolor[HTML]{FFFFFF}\textbf{AUC}} & \multicolumn{2}{c}{\cellcolor[HTML]{FFFFFF}\textbf{F1}} & \textbf{P@N} \\ \cline{2-3} \cline{4-5} \cline{6-6}
\rowcolor[HTML]{FFFFFF} 
\multirow{-2}{*}{\cellcolor[HTML]{FFFFFF}\textbf{}}        & \textbf{Macro}              & \textbf{Micro}             & \textbf{Macro}            & \textbf{Micro}              & \textbf{P@5}                  \\ \hline
\rowcolor[HTML]{F3F3F3} 
\textbf{M = 1}                                             & $93.5$                        & $95.2$                       & $69.3$                       & $72.5$                       & $68.0$                 \\
\textbf{M = 2 (random)}                                    & $93.3$                        & $95.2$                       & $69.4$                       & $72.8$                       & $68.2$                 \\
\textbf{M = 2 (maximum-diversity)}                         & $93.6$                        & $95.3$                       & $69.8$                       & $73.4$                       & $68.3$                 \\
\rowcolor[HTML]{F3F3F3} 
\textbf{M = 4 (random)}                                    & $93.6$                        & $95.3$                       & $69.8$                       & $73.3$                       & $68.4$                 \\
\rowcolor[HTML]{F3F3F3} 
\cellcolor[HTML]{F3F3F3}\textbf{M = 4 (maximum-diversity)} & \textbf{93.7}                        & \textbf{95.4}                      & \textbf{70.4}                      & \textbf{73.9}                       & \textbf{68.8}                 \\
\textbf{M = 8 (random)}                                    & $93.4$                        & $95.1$                       & $69.9$                       & $73.4$                       & $68.2$                 \\
\textbf{M = 8 (maximum-diversity)}                         & $93.4$                        & $95.1$                       & $69.2$                       & $72.9$                       & $68.0$ \\ \hline          
\end{tabular}
}
\caption{Results when considering a different number of synonyms ($M$) on the MIMIC-III-50 dataset.}
\label{table:M-settings}
\end{table}

\begin{table}[t!]
\renewcommand{\arraystretch}{1.25}
\centering
\resizebox{\columnwidth}{!}{%
\begin{tabular}{llcccc}
\hline
\rowcolor[HTML]{FFFFFF} 
\textbf{Dataset}                  & \textbf{Classifier} & \textbf{Mean} & \textbf{Low Pth} & \cellcolor[HTML]{FFFFFF}\textbf{Medium Pth} & \cellcolor[HTML]{FFFFFF}\textbf{High Pth} \\ \hline
\rowcolor[HTML]{F3F3F3} 
\textbf{}                         & BM                  & $3.8$e-$02$           & \textbf{1.7e-02}            & $3.0$e-$02$                        & $6.1$e-$02$                         \\
\cellcolor[HTML]{F3F3F3}\textbf{} & LE                  & $6.0$e-$02$           & $3.5$e-$02$                 & $6.4$e-$02$                        & $8.2$e-$02$                         \\
\cellcolor[HTML]{F3F3F3}\textbf{} & CE                  & $3.5$e-$02$           & $2.1$e-$02$                 & $3.0$e-$02$                        & $5.1$e-$02$                         \\
\rowcolor[HTML]{F3F3F3}
\textbf{MIMIC-III-50}             & LE+MSAM             & $2.7$e-$02$           & $2.0$e-$02$                 & $2.5$e-$02$                        & $3.6$e-$02$                         \\
\rowcolor[HTML]{F3F3F3} 
                                  & CE+MSAM             & $2.8$e-$02$           & $1.9$e-$02$                 & \textbf{2.4e-02}                   & $3.7$e-$02$                         \\
\cellcolor[HTML]{F3F3F3}          & CE+MSAM+$\text{CLQ}_\text{MSE}$                   & $2.9$e-$02$                 & $2.1$e-$02$                        & $2.6$e-$02$                        & $3.7$e-$02$                         \\
\cellcolor[HTML]{F3F3F3}          & CE+MSAM+$\text{CLQ}_\text{Huber}$       & \textbf{2.6e-02}  & $1.9$e-$02$                     & $2.5$e-$02$                        & \textbf{3.3e-02}        \\ \hline
\rowcolor[HTML]{F3F3F3} 
\textbf{}                         & BM                  & $218.1$e-$05$           & \textbf{8.8e-05}            & \textbf{60.0e-05}            & $1520.0$e-$05$                         \\
\cellcolor[HTML]{F3F3F3}\textbf{} & LE                  & $258.2$e-$05$           & $12.2$e-$05$                     & $90.0$e-$05$                        & $1660.8$e-$05$                         \\
\cellcolor[HTML]{F3F3F3}\textbf{} & CE                  & $248.6$e-$05$           & $11.7$e-$05$                    & $87.8$e-$05$                     & $1595.1$e-$05$                         \\
\rowcolor[HTML]{F3F3F3} 
\textbf{MIMIC-III-clean}          & LE+MSAM            & \textbf{140.1e-05}  & $18.4$e-$05$                    & $81.2$e-$05$                     & \textbf{680.1e-05}                         \\
\rowcolor[HTML]{F3F3F3} 
\textbf{}                         & CE+MSAM             & $155.6$e-$05$           & $18.5$e-$05$                    & $85.3$e-$05$                     & $769.0$e-$05$                     \\
\cellcolor[HTML]{F3F3F3}\textbf{} & CE+MSAM+$\text{CLQ}_\text{MSE}$         & $161.4$e-$05$           & $20.1$e-$05$                    & $87.7$e-$05$                     & $800.6$e-$05$                        \\ 
\cellcolor[HTML]{F3F3F3}\textbf{} & CE+MSAM+$\text{CLQ}_\text{Huber}$         & $157.2$e-$05$           & $18.3$e-$05$                    & $85.3$e-$05$                     & $780.6$e-$05$        \\ \hline
\end{tabular}
}
\caption{Calibration quality according to the MECE metric, for all the proposed classification models and on different percentiles of the MIMIC-III splits.}
\label{table:MECE}
\end{table}

The results show that the label embedding mechanism offers notable benefits in model calibration. The joint optimization also improved calibration results for MIMIC-III-50, although not in MIMIC-III-clean, where LE+MSAM exhibited the best average calibration performance. Chunk-based modeling can perhaps overvalue the probabilities associated to each class, negatively affecting the calibration with its use of a max-pooling operation. Additionally, our results also show that the Huber loss has a positive effect on enhancing model calibration, compared to the MSE loss. Although the jointly trained model that optimizes quantification through the MSE loss has slightly better classification results, the variant that uses the Huber loss is better in terms of calibration. We, therefore, used the CE+MSAM+$\text{CLQ}_\text{Huber}$ model for further considerations, arguing that this variant can better balance classification and calibration performance.

Compared to other approaches in the literature, we outperform the previously best-performing models reported for the two MIMIC-III splits under analysis. It is worth noting that the models reported by \citet{Edin23} underwent an adjustment using the validation splits, as the authors reported on classification performance after optimizing the decision boundary values through a grid search mechanism to maximize F1 scores in the validation splits. In contrast, our results do not involve any such adjustment and still surpass the best-reported models to date, establishing a new state-of-the-art with a default decision boundary set at $0.5$. 

On MIMIC-III-50, the proposed approach outperforms the best reported model to date (i.e., KEPTLongFormer) across all metrics, with scores of $93.7$ (+$1.1$), $95.4$ (+$0.6$), $70.3$ (+$1.4$), $73.8$ (+$0.9$), and $68.9$ (+$1.6$) in terms of macro-AUC, micro-AUC, macro-F1, micro-F1, and P@5, respectively. On MIMIC-III-clean, we outperform the best reported model to date (i.e., PLM-ICD) also across all metrics, with scores of $96.3$ (+$0.4$), $98.9$ (+$0.0$), $30.5$ (+$3.9$), $60.4$ (+$0.8$), $73.3$ (+$1.2$) and $57.4$ (+$0.9$) in terms of macro-AUC, micro-AUC, macro-F1, micro-F1, P@8, and P@15. 

Appendix \ref{appendix: classification} details the classification performance across different ICD chapters, additionally also showing results for the top-10 most frequent ICD codes, and for relevant chronic diseases. These different examples attest to the usefulness of our approach, offering accurate classification results that can inform different types of downstream analyses.

\subsubsection{Quantification}

Tables \ref{table:quantification-50} and \ref{table:quantification-clean} show results for quantification experiments using both MIMIC-III splits. The baseline results correspond to the standard CC and PCC methods, and also to the use of an MLP separately trained for quantification, following the experimental setup from \citet{coutinho2023exploring}. In the case of the proposed models, i.e. CE+MSAM+$\text{CLQ}_\text{MSE}$ and CE+MSAM+$\text{CLQ}_\text{Huber}$, the MLP trained jointly with the classifier, using either the MSE or the Huber loss, was employed for quantification.

CE+MSAM+$\text{CLQ}_\text{MSE}$ outperforms CE+MSAM and CE+MSAM+$\text{CLQ}_\text{Huber}$ in terms of classification accuracy, but not in terms of calibration, which impacts the standard CC and PCC metrics. Notably, the LE+MSAM model achieves better results with PCC but worse with CC, aligning with the idea that chunk encoding tends to overvalue class probabilities, negatively impacting the PCC results. Still, this effect is mitigated when not looking at the probabilities directly, i.e. with the CC method.

Analyzing Tables \ref{table:quantification-50} and \ref{table:quantification-clean}, we see that joint optimization generally surpasses all mentioned baselines, including the separate training of the MLP proposed by \citet{coutinho2023exploring}. This is consistent across all metrics, except for the MRAE in MIMIC-III-clean, where combining BM with PCC corresponds to the lowest value. Although this result seems misleading, the MRAE uses additive smoothing to avoid divisions by zero, which slightly impacts results according to group sizes. Is is thus important to also look at results with the MAE, i.e., a complementary metric that evaluates quantification consistently across group sizes.

Overall, the results indicate that joint training, and particularly when using the Huber loss, can effectively inform the MLP about class distributions, leading to good quantification performance and improving classifier calibration compared to the MSE loss. We present a more detailed analysis of the quantification results in Appendix \ref{appendix: quantification}.

\begin{table}[t!]
\renewcommand{\arraystretch}{1.25}
\centering
\resizebox{\columnwidth}{!}{%
\begin{tabular}{lcccccc}
\hline
\rowcolor[HTML]{FFFFFF} 
\cellcolor[HTML]{FFFFFF}                                 & \multicolumn{2}{c}{\cellcolor[HTML]{FFFFFF}\textbf{CC}} & \multicolumn{2}{c}{\cellcolor[HTML]{FFFFFF}\textbf{PCC}} & \multicolumn{2}{c}{\cellcolor[HTML]{FFFFFF}\textbf{MLP/CLQ}} \\ \cline{2-7} 
\rowcolor[HTML]{FFFFFF} 
\multirow{-2}{*}{\cellcolor[HTML]{FFFFFF}\textbf{Model}} & \textbf{MAE}               & \textbf{MRAE}              & \textbf{MAE}               & \textbf{MRAE}               & \textbf{MAE}                 & \textbf{MRAE}                 \\ \hline
\rowcolor[HTML]{F3F3F3} 
BM                                                       & $4.08$e-$02$          & $22.02$e-$02$          & $1.70$e-$02$          & $~10.12$e-$02$         & $1.22$e-$02$         & $7.22$e-$02$              \\
LE                                                       & $2.83$e-$02$          & $15.70$e-$02$          & $2.02$e-$02$          & $11.53$e-$02$          & $1.15$e-$02$         & $6.87$e-$02$     \\
CE                                                       & $2.11$e-$02$          & $10.08$e-$02$          & $1.50$e-$02$          & $~9.67$e-$02$          & $1.14$e-$02$         & $6.83$e-$02$              \\
\rowcolor[HTML]{F3F3F3} 
LE+MSAM                                                  & $1.96$e-$02$          & $10.11$e-$02$          & $1.25$e-$02$          & $~7.34$e-$02$          & $1.09$e-$02$         & \textbf{6.64e-02}     \\
\rowcolor[HTML]{F3F3F3} 
CE+MSAM                                                  & $1.72$e-$02$          & $9.31$e-$02$           & $1.38$e-$02$          & $~9.31$e-$02$         & $1.09$e-$02$         & \textbf{6.64e-02}     \\
CE+MSAM+$\text{CLQ}_\text{MSE}$                                              & $1.91$e-$02$          & $9.90$e-$02$           & $1.69$e-$02$          & $10.90$e-$02$          & $1.14$e-$02$         & $6.83$e-$02$        \\
CE+MSAM+$\text{CLQ}_\text{Huber}$         & $1.89$e-$02$          & $10.32$e-$02$          & $1.47$e-$02$          & $10.0$e-$02$           & \textbf{1.05e-02}     & \textbf{6.64e-02}        \\ \hline
\end{tabular}
}
\caption{Results for different quantification methods, using the results from different classification models on the MIMIC-III-50 test dataset split.}
\label{table:quantification-50}
\end{table}

\begin{table}[t!]
\renewcommand{\arraystretch}{1.25}
\centering
\resizebox{\columnwidth}{!}{%
\begin{tabular}{lcccccc}
\hline
\rowcolor[HTML]{FFFFFF} 
\cellcolor[HTML]{FFFFFF}                                 & \multicolumn{2}{c}{\cellcolor[HTML]{FFFFFF}\textbf{CC}} & \multicolumn{2}{c}{\cellcolor[HTML]{FFFFFF}\textbf{PCC}} & \multicolumn{2}{c}{\cellcolor[HTML]{FFFFFF}\textbf{MLP/CLQ}} \\ \cline{2-7} 
\rowcolor[HTML]{FFFFFF} 
\multirow{-2}{*}{\cellcolor[HTML]{FFFFFF}\textbf{Model}} & \textbf{MAE}               & \textbf{MRAE}              & \textbf{MAE}               & \textbf{MRAE}               & \textbf{MAE}                 & \textbf{MRAE}                 \\ \hline
\rowcolor[HTML]{F3F3F3} 
BM                                                       & $18.73$e-$04$          & $3.03$e-$01$       & $8.39$e-$04$       & \textbf{1.88e-01}       & $8.07$e-$04$         &   $2.18$e-$01$            \\
LE                                                       & $16.00$e-$04$          & $2.86$e-$01$       & $9.99$e-$04$       & $2.01$e-$01$       & $8.00$e-$04$         & $2.18$e-$01$     \\
CE                                                       & $15.78$e-$04$          & $2.76$e-$01$       & $11.23$e-$04$       & $2.02$e-$01$       & $7.89$e-$04$       & $2.12$e-$01$              \\
\rowcolor[HTML]{F3F3F3} 
LE+MSAM                                                  & $12.14$e-$04$          & $2.84$e-$01$       & $7.34$e-$04$       & $2.21$e-$01$      & $7.83$e-$04$         & $2.13$e-$01$     \\
\rowcolor[HTML]{F3F3F3} 
CE+MSAM                                                  & $11.48$e-$04$          & $2.27$e-$01$       & $7.78$e-$04$       & $2.21$e-$01$      & $7.68$e-$04$      &   $2.09$e-$01$             \\
CE+MSAM+$\text{CLQ}_\text{MSE}$                          & $11.13$e-$04$          & $2.26$e-$01$       & $9.12$e-$04$       & $2.55$e-$01$      & $7.00$e-$04$       & $2.04$e-$01$        \\
CE+MSAM+$\text{CLQ}_\text{Huber}$                        & $11.41$e-$04$          & $2.28$e-$01$       & $8.12$e-$04$       & $2.30$e-$01$      & \textbf{6.93e-04}         & $2.04$e-$01$  \\ \hline
\end{tabular}
}
\caption{Results for different quantification methods, using the results from different classification models on the MIMIC-III-clean test dataset split.}
\label{table:quantification-clean}
\end{table}

\section{Conclusions and Future Work}

This work introduced a novel deep learning method for ICD coding, achieving state-of-the-art results in tests with two well established MIMIC-III dataset splits. The proposed method processes long clinical documents and uses a label embedding mechanism that explores diverse ICD code synonyms. Besides achieving highly accurate classification results, the proposed approach produces well-calibrated estimates that can effectively inform downstream tasks such as text quantification.

Despite the strong results, it should be noted that our model does not exploit the hierarchical structure inherent to the ICD coding system. Thus, a promising avenue for further improvement involves using this structural knowledge, e.g. by implementing dual classification heads. Another path worth exploring relates to using alternative methods to improve calibration (e.g., using other classification loss functions besides the BCE), since improving calibration is beneficial for classification and essential for accurate results in quantification.

\section*{Acknowledgements}

We thank the anonymous reviewers for their valuable comments and suggestions. This research was supported by the Portuguese Recovery and Resilience Plan through project C645008882-00000055 (i.e., the Center For Responsible AI), and also by Funda\c{c}\~ao para a Ci\^encia e Tecnologia (FCT), through the projects with references DSAIPA/DS/0133/2020 and UIDB/50021/2020 (DOI:10.54499/UIDB/50021/2020).

\section*{Limitations and Ethical Considerations}

While our work does not raise new ethical issues within this domain, there are general concerns. 

ICD coding is essential in clinical, operational, and financial healthcare decisions. Traditionally, medical coders review documents and manually assign the appropriate ICD codes by following specific coding guidelines. Approaches such as ours can significantly reduce time and costs in ICD coding. Still, there are risks associated with over-reliance on automatic coding methods. No matter how accurate a given approach is, it is still possible to misclassify documents with erroneous ICD codes, affecting patient treatment. Therefore, automatic coding should assist, rather than replace, the judgment of trained clinical professionals.

Our experiments have also relied on MIMIC-III dataset splits used in previous studies. While these datasets constitute useful benchmarks for developing and evaluating new methods, they do not represent the enormous variety of clinical and linguistic data encountered in potential deployments.
\bibliography{custom}
\bibliographystyle{acl_natbib}

\clearpage
\appendix
\section{Appendix}
\label{appendix}
This appendix presents statistical information about the dataset splits, and additional experimental results for the classification and quantification tasks.

\subsection{MIMIC-III Dataset Splits}
\label{appendix: dataset}

Table \ref{table:splits} provides a statistical characterization of the MIMIC-III splits considered in our experiments, detailing the training, validation, and test sets, and underlining the highly imbalanced label distribution, the disparity between the average and maximum document lengths, and the high number of ICD codes assigned to each discharge summary.
\begin{table}[b!]
\renewcommand{\arraystretch}{1.54}
\centering
\resizebox{\columnwidth}{!}{%
\begin{tabular}{lcccccccccc}
\hline
\rowcolor[HTML]{FFFFFF} 
\cellcolor[HTML]{FFFFFF}                                       & \cellcolor[HTML]{FFFFFF}                                   & \multicolumn{2}{c}{\cellcolor[HTML]{FFFFFF}\textbf{Words per Doc.}} & \multicolumn{2}{c}{\cellcolor[HTML]{FFFFFF}\textbf{Tokens per Doc.}} & \multicolumn{2}{c}{\cellcolor[HTML]{FFFFFF}\textbf{Codes per Doc.}} & \textbf{Unique}          & \multicolumn{2}{c}{\cellcolor[HTML]{FFFFFF}\textbf{Type of Codes}} \\ \cline{3-8} \cline{10-11} 
\rowcolor[HTML]{FFFFFF} 
\multirow{-2}{*}{\cellcolor[HTML]{FFFFFF}\textbf{Set (Split)}} & \multirow{-2}{*}{\cellcolor[HTML]{FFFFFF}\textbf{Samples}} & \textbf{Avg.}                    & \textbf{Max.}                    & \textbf{Avg.}                     & \textbf{Max.}                    & \textbf{Avg.}                    & \textbf{Max.}                    & \textbf{Codes}           & \textbf{Diag.}                   & \textbf{Proc.}                  \\ \hline
\rowcolor[HTML]{F3F3F3} 
Train (top-50)            &$8,066$ & $1,642$ & $7,989$ & $2,830$ & $20,297$ & $5.4$ & $18$ & $50$ & $33$ & $17$                  \\
\cellcolor[HTML]{F3F3F3}Val. (top-50)          & $1,573$ & $1,932$ & $6,658$ & $3,410$ & $16,566$ & $5.9$ & $21$ & $50$ & $33$ & $17$                  \\
\rowcolor[HTML]{F3F3F3} 
Test (top-50)              & $1,729$ & $1,964$ & $6,470$ & $3,465$ & $11,871$ & $6.0$ & $20$ & $50$ & $33$ & $17$                  \\
Train (clean)              &$38,401$ & $1,514$ & $10,500$ & $1,651$ & $11,758$ & $14.0$ & $57$ & $3,681$ & $2,849$ & $832$                 \\
Val. (clean)                                                   & \cellcolor[HTML]{F3F3F3}$5,577$                                  & \cellcolor[HTML]{F3F3F3}$1,552$       & \cellcolor[HTML]{F3F3F3}$6,393$     & \cellcolor[HTML]{F3F3F3}$1,694$       & \cellcolor[HTML]{F3F3F3}$6,897$       & \cellcolor[HTML]{F3F3F3}$15.9$      & \cellcolor[HTML]{F3F3F3}$60$      & \cellcolor[HTML]{F3F3F3}$3,676$ & \cellcolor[HTML]{F3F3F3}$2,844$       & \cellcolor[HTML]{F3F3F3}$832$      \\
Test (clean)                & $8,734$ & $1,485$ & $7,858$ & $1,619$ & $8,299$ & $14.8$ & $56$ & $3,681$ & $2,849$ & $832$                 \\\hline
\end{tabular}
}
\caption{Statistics for the training, validation and test sets of MIMIC-III-50 (top) and MIMIC-III-clean (bottom). The columns labeled with {\it words per doc} refer to the average and maximum number of words per hospital discharge summary. {\it Tokens per doc} corresponds to the average and maximum number of tokens per clinical document. {\it Codes per doc} refers to the average and maximum number of ICD codes per document. {\it Unique codes} corresponds to the number of distinct ICD codes. Finally, {\it type of codes} is used to indicate the number of distinct diagnosis and procedure codes.}
\label{table:splits}
\end{table}

\begin{table}[b!]
\renewcommand{\arraystretch}{1.54}
\centering
\resizebox{0.8\columnwidth}{!}{%
\begin{tabular}{llccc}
\hline
\rowcolor[HTML]{FFFFFF} 
\textbf{Dataset}                                                   & \textbf{Split} & \textbf{Low Pth} & \textbf{Medium Pth} & \textbf{High Pth} \\ \hline
\rowcolor[HTML]{F3F3F3} 
\cellcolor[HTML]{F3F3F3}                                  & Train & 397-449           & 759-914              & 1615-3233             \\
\multirow{-2}{*}{\cellcolor[HTML]{F3F3F3}\textbf{MIMIC-III-50}}    & Test  & 60-127            & 148-247              & 402-470               \\
\rowcolor[HTML]{F3F3F3} 
\cellcolor[HTML]{F3F3F3}                                  & Train & 4-9               & 36-56                & 308-14,598            \\
\multirow{-2}{*}{\cellcolor[HTML]{F3F3F3}\textbf{MIMIC-III-clean}} & Test  & 1-4               & 6-27                 & 55-2228               \\ \hline
\end{tabular}
}
\caption{Number of ICD code occurrences in specific percentiles of code prevalence frequency.}
\label{table: freq_interval}
\end{table}

In turn, Table \ref{table: freq_interval} presents the frequency of ICD codes over the data, dividing the ICD codes into three relevant percentiles for the training and test sets, for both MIMIC-III splits. Low Pth accounts for the 10\% of medical codes with the lowest occurrence frequency in the training set of the respective MIMIC-III split. Medium Pth corresponds to the 10\% of codes with medium occurrence frequency, falling within the 55\% to 65\% percentile range in the training set of the respective MIMIC-III split. Lastly, High Pth corresponds to the 10\% of codes with the highest occurrence frequency in the training set of the respective MIMIC-III split.

\subsection{Additional Classification Results}
\label{appendix: classification}

Tables \ref{table:top10} to \ref{table:procedure_chpaters} provide additional insights into our classification results, specifically considering the CE+MSAM+$\text{CLQ}_\text{Huber}$ model.

Table \ref{table:top10} presents classification results for the top-10 most frequent ICD codes over MIMIC-III-clean. We obtained a mean precision of $76.46\%$, a mean recall of $80.8\%$, and a mean F1 score of $78.50\%$. In turn, Table \ref{table:chronic_diseases} presents the performance for examples of relevant chronic diseases, representing some of the main focuses of healthcare investigation. Tables \ref{table:diagnosis_chpaters} and \ref{table:procedure_chpaters} provide results for codes within different ICD diagnosis and procedure chapters. Together, these results illustrate different possible applications for the ICD coding results.

\begin{table}[t!]
\renewcommand{\arraystretch}{1.54}
\centering
\resizebox{\columnwidth}{!}{%
\begin{tabular}{llccc}
\hline
\rowcolor[HTML]{FFFFFF} 
\textbf{Code}                                    & \textbf{Description}                                             & \textbf{Precision}                         & \textbf{Recall}                            & \textbf{F1}                                \\ \hline
\rowcolor[HTML]{F3F3F3} 
401.9                                            & \textit{Unspecified essential hypertension}                       & $76.73$                  & $85.20$                  & $80.75$                  \\
38.93                                            & \textit{Venous Catheterization, Not Elsewhere Classified}        & $69.30$                  & $71.72$                  & $70.49$                  \\
\rowcolor[HTML]{F3F3F3} 
428.0                                            & \textit{Heart failure}                                           & $80.94$                  & $82.66$                  & $81.79$                  \\
427.31                                           & \textit{Atrial fibrillation}                                     & $90.57$                  & $91.82$                  & $91.19$                  \\
\rowcolor[HTML]{F3F3F3} 
414.01                                           & \textit{Coronary atherosclerosis of native coronary artery}      & $81.17$                  & $86.35$                  & $83.68$                  \\
96.04                                            & \textit{Insertion Of Endotracheal Tube}                          & $79.78$                  & $80.99$                  & $80.38$                  \\
\rowcolor[HTML]{F3F3F3} 
96.6                                             & \textit{Enteral Infusion Of Concentrated Nutritional Substances} & $70.24$                  & $76.86$                  & $73.40$                 \\
584.9                                            & \textit{Acute kidney failure, unspecified}                       & $73.70$                  & $68.79$                  & $71.16$                 \\
\rowcolor[HTML]{F3F3F3} 
\cellcolor[HTML]{F3F3F3}                         & \textit{Diabetes mellitus without mention of complication}       & \cellcolor[HTML]{F3F3F3}                        & \cellcolor[HTML]{F3F3F3}                        & \cellcolor[HTML]{F3F3F3}                          \\
\rowcolor[HTML]{F3F3F3} 
\multirow{-2}{*}{\cellcolor[HTML]{F3F3F3}250.00} & \textit{type II or unspecified type, not stated as uncontrolled} & \multirow{-2}{*}{\cellcolor[HTML]{F3F3F3}72.43} & \multirow{-2}{*}{\cellcolor[HTML]{F3F3F3}83.37} & \multirow{-2}{*}{\cellcolor[HTML]{F3F3F3}77.52} \\
272.4                                            & \textit{Other and unspecified hyperlipidemia}                    & $69.73$                  & $80.24$                  & $74.62$                  \\ \hline
\rowcolor[HTML]{FFFFFF} 
\multicolumn{2}{c}{\cellcolor[HTML]{FFFFFF}\textbf{Average}}                                                        & $76.46$                  & $80.8$                & $78.50$                \\ \hline
\end{tabular}
}
\caption{Results over the test split for the 10 most frequent ICD codes in the MIMIC-III-clean dataset.}
\label{table:top10}
\end{table}

\begin{table}[t!]
\renewcommand{\arraystretch}{1.54}
\centering
\resizebox{\columnwidth}{!}{%
\begin{tabular}{llcccc}
\hline
\rowcolor[HTML]{FFFFFF} 
\cellcolor[HTML]{FFFFFF}                                 & \cellcolor[HTML]{FFFFFF}                                           & \textbf{Unique codes} & \cellcolor[HTML]{FFFFFF}                                      & \multicolumn{2}{c}{\cellcolor[HTML]{FFFFFF}\textbf{Performance Metrics}} \\ \cline{5-6} 
\rowcolor[HTML]{FFFFFF} 
\multirow{-2}{*}{\cellcolor[HTML]{FFFFFF}\textbf{Block}} & \multirow{-2}{*}{\cellcolor[HTML]{FFFFFF}\textbf{Chronic Disease}} & \textbf{(Present)}    & \multirow{-2}{*}{\cellcolor[HTML]{FFFFFF}\textbf{Percentage}} & \textbf{Macro-F1}                   & \textbf{Micro-F1}                  \\ \hline
\rowcolor[HTML]{F3F3F3}
250                             & \textit{Diabetes mellitus}         & $33$ & $1.943\%$ & $28.60$ & $65.22$                 \\
401-405                         & \textit{Hypertensive Disease}      & $14$ & $3.303\%$ & $29.10$ & $76.84$                 \\
\rowcolor[HTML]{F3F3F3}
410-414                         & \textit{Ischemic Heart Disease}    & $32$ & $3.279\%$ & $29.49$ & $69.01$                  \\
428                             & \textit{Heart Failure}             & $15$ & $2.471\%$ & $36.65$ & $71.47$                 \\
\rowcolor[HTML]{F3F3F3}
585;403-404                     & \textit{Renal Failure}             & $16$  & $1.600\%$ & $34.95$ & $58.71$                  \\
490-496                         & \textit{Pulmonary Disease}         & $16$ &$ 1.209\%$ & $45.05$ & $67.56$               \\ \hline
\end{tabular}
}
\caption{Results for relevant chronic diseases. The columns {\it unique codes} and {\it percentage} refer to the number of unique codes of the respective block within the MIMIC-III-clean dataset, and to the corresponding percentage of occurrences over the dataset.}
\label{table:chronic_diseases}
\end{table}

\begin{table}[t!]
\renewcommand{\arraystretch}{1.54}
\centering
\resizebox{\columnwidth}{!}{%
\begin{tabular}{lcccccccccc}

\hline
\rowcolor[HTML]{FFFFFF} 
\textbf{}        &  & \multicolumn{3}{c}{\cellcolor[HTML]{FFFFFF}\textbf{Occurrences}} &  & \textbf{}           &  & \multicolumn{2}{c}{\cellcolor[HTML]{FFFFFF}\textbf{Preformance Metrics}} \\ \cline{3-5} \cline{9-10} 
\rowcolor[HTML]{FFFFFF} 
\textbf{Chapter} &  & \textbf{Train}     & \textbf{Validation}     & \textbf{Test}     &  & \textbf{Percentage} &  & \textbf{Macro-F1}                   & \textbf{Macro-F1}                  \\ \hline
\rowcolor[HTML]{F3F3F3}
 I & & $14,050$ & $2,090$ & $3,212$ & & $3.190$\% & & $33.57$ & $52.91$ \\

 II & & $9,200$ &$ 1,401 $&$ 2,076 $& & $2.09\%$ & & $36.35$ & $57.62$ \\
\rowcolor[HTML]{F3F3F3}
 III & & $49,135$ & $7,356$ & $11,008 $& & $11.126\%$ & & $32.99$ & $60.33$\\

 IV & & $17,882$ & $2,657$ & $4,106 $& & $4.062\%$ & & $29.94$ & $40.93$ \\ 
\rowcolor[HTML]{F3F3F3}
 V & & $17,392 $& $2,562$ &$ 3,740$ & & $3.905\%$ & & $21.87$ & $47.90$ \\

 VI & & $15,811$	& $2,433$ 	& $3,397$	& & $3.567\%$ & & $28.78$ & $54.69$\\
\rowcolor[HTML]{F3F3F3}
 VII & & $99,076$ & $14,729$ & $22,526$ & & $22.471\%$ & & $29.08$ & $67.38$\\

 VIII & & $31,613$ & $4,703$ & $7,113$ & & $7.158\%$ & & $35.46$ & $95.54$ \\
\rowcolor[HTML]{F3F3F3}
 IX & & $27,061$ & $3,967$ & $ 6,022$ & & $6.107\%$ & & $30.59$ & $56.32$ \\

 X & & $22,940$ & $3,438$ & $5,260$ & & $5.215\%$ & & $28.90$ & $62.00$ \\
\rowcolor[HTML]{F3F3F3}
 XI & & $151$ & $24$ & $33$ & & $0.034\%$ & & $25.95$ & $40.00$\\

 XII & & $6,056$ & $888$ & $1,371$ & & $1.371\%$ & & $28.64$ & $47.17$\\
\rowcolor[HTML]{F3F3F3}
 XIII & & $9,098$ & $1,360$ & $1,944$ & & $2.044\%$ & & $27.66$ & $51.08$ \\

 XIV & & $2,228$ & $328$ & $471$ & & $0.499\%$ & & $51.85$ & $62.61$ \\
\rowcolor[HTML]{F3F3F3}
 XV	& & $12,656$ & $1,740$ & $2,565$ & & $2.796\%$ & & $31.18$ & $59.92$ \\

 XVI & & $20,692$ & $3,154$ & $4,550$ & & $4.68\%$ & & $15.32$ & $39.50$ \\
\rowcolor[HTML]{F3F3F3}
 XVII & & $87,280$ & $13,018$ & $19,131$ & & $19.685\%$ & & $23.56$ & $50.83$ \\
\hline
\end{tabular}
}
\caption{Number of instances and classification performance metrics for each of the ICD diagnosis chapters. The column named {\it percentage} corresponds to the percentage of the diagnosis codes under consideration over the MIMIC-III-clean dataset.}
\label{table:diagnosis_chpaters}
\end{table}

\begin{table}[t!]
\renewcommand{\arraystretch}{1.54}
\centering
\resizebox{\columnwidth}{!}{%
\begin{tabular}{lcccccccccc}

\hline
\rowcolor[HTML]{FFFFFF} 
\textbf{}        &  & \multicolumn{3}{c}{\cellcolor[HTML]{FFFFFF}\textbf{Occurrences}} &  & \textbf{}           &  & \multicolumn{2}{c}{\cellcolor[HTML]{FFFFFF}\textbf{Preformance Metrics}} \\ \cline{3-5} \cline{9-10} 
\rowcolor[HTML]{FFFFFF} 
\textbf{Chapter} &  & \textbf{Train}     & \textbf{Validation}     & \textbf{Test}     &  & \textbf{Percentage} &  & \textbf{Macro-F1}                   & \textbf{Macro-F1}                  \\ \hline
\rowcolor[HTML]{F3F3F3}
 I & & $5,508$ & $855$ & $1,347$ & & $3.589\%$ & & $36.48$ & $65.02$ \\

 II & & $4,852$ & $733$ & $1,148$ & & $3.134\%$ & & $41.18$ & $66.73$ \\
\rowcolor[HTML]{F3F3F3}
 III & & $91$ & $13$ & $17$ & & $0.056\%$ & & $63.07$ & $66.67$ \\

 IV	& & $102$ & $15$ & $23$ & & $0.065\%$ & & $56.89$ & $59.57$ \\
\rowcolor[HTML]{F3F3F3}
 V & & $0$ & $0$ & $0$ & & $0\%$ & & $0.0$ & $0.0$\\

 VI	& & $21$ & $3$ & $4$ & & $0.013\%$ & & $40.00$ & $40.00$ \\
\rowcolor[HTML]{F3F3F3}
 VII & & $501$ & $75$ & $104$ & & $0.317\%$ & & $24.58$ & $36.59$ \\

 VIII & & $9,590$ & $1,480$ & $2,164$ & & $6.161\%$ & & $37.12$ & $63.90$ \\
\rowcolor[HTML]{F3F3F3}
 IX	& & $47,762$ & $6,895$ & $10,813$ & & $30.478\%$ & & $46.07$ & $76.23$ \\

 X & & $897$ & $127$ & $217$ & & $0.578\%$ & & $49.38$ & $71.36$ \\
\rowcolor[HTML]{F3F3F3}
 XI & & $15,302$ & $2,267$ & $3,555$ & & $9.834\%$ & & $38.18$ & $66.54$ \\

 XII & & $1,045$ & $152$ & $230$ & & $0.664\%$ & & $54.41$ & $74.89$ \\
\rowcolor[HTML]{F3F3F3}
 XIII & & $641$ & $102$ & $127$ & & $0.405\%$ & & $74.62$ & $69.96$ \\

 XIV & & $201$ & $27$ & $43$ & & $0.126\%$ & & $56.65$ & $65.06$\\
\rowcolor[HTML]{F3F3F3}
 XV	& & $20$ & $3$ & $4$ & & $0.013\%$ & & $88.89$ & $88.89$ \\

 XVI & &  $5,990$ & $924$ & $1,307$ & & $3.827\%$ & & $42.50$ & $59.86$ \\
\rowcolor[HTML]{F3F3F3}
 XVII & & $2,308$ & $318$ & $539$ & & $1.473\%$ & & $32.19$ & $50.41$ \\

 XVIII & & $61,329$ & $8,568$ & $14,455$ & & $39.267\%$ & & $25.37$ & $66.70$ \\
\hline
\end{tabular}
}
\caption{Number of instances and classification performance metrics for each of the ICD procedure chapters. The column named {\it percentage} corresponds to the percentage of the procedure codes under consideration over the MIMIC-III-clean dataset.}
\label{table:procedure_chpaters}
\end{table}

Note that Chapter VII (i.e., \textit{diseases of the circulatory system}) in the ICD-9 diagnosis codes accounts for a substantial portion of the MIMIC-III-clean dataset, representing $22.471\%$ of all diagnosis codes. This chapter demonstrates good classification performance, with our model achieving a macro-averaged F1 score of $29.08\%$ and a micro-averaged F1 score of $67.38\%$.

Conversely, Chapter XI (i.e., \textit{complications of pregnancy, childbirth, and the puerperium}) is the least frequent chapter of ICD codes, and corresponds to the lowest classification performance. With a prevalence of only $0.034\%$ in the dataset, our model achieved macro- and micro-averaged F1 scores of $25.95\%$ and $40.00\%$, respectively in this chapter. These scores highlight the negative impact of infrequent ICD codes on model effectiveness. 

Furthermore, we observe an interesting phenomenon in Chapter XIV (i.e., \textit{congenital anomalies}). Although this chapter represents a relatively small percentage ($0.499\%$) of the overall dataset, the model performs remarkably well in this chapter. It attains macro and micro-averaged F1 scores of $51.85\%$ and $62.61\%$, respectively, empirically showing the model's ability to perform few-shot learning when dealing with seldom-seen codes.

When we examine the overall distribution of procedure codes, we see that the dataset is characterized by a generally low density of procedure codes, with two notable exceptions in Chapter IX (i.e., \textit{operations on the cardiovascular system}) and Chapter XVIII (i.e., \textit{miscellaneous diagnostic and therapeutic procedures}), which encompass almost $70\%$ of the dataset. However, despite the relatively low frequency of procedures in the other chapters, our model performs exceptionally well in them. For instance, Chapters VI and XV achieve performance values of $40\%$ and $88.89\%$ respectively in macro- and micro-averaged F1, even though these codes have a small representation of $0.013\%$ within the dataset. These results highlight the model's capacity to learn even from infrequent instances, again emphasizing its few-shot learning capabilities.

Chapter XVIII in the ICD-9 procedure codes, which covers \textit{miscellaneous diagnostic and therapeutic procedures}, stands out as the most frequently occurring chapter in the MIMIC-III-clean dataset, accounting for a substantial $39.267\%$ of the total number of instances. In this chapter, we achieve $25.37\%$ for macro-averaged F1 and $66.70\%$ for micro-averaged F1.

\subsection{Additional Quantification Results}
\label{appendix: quantification}

\begin{figure}[t!]
  \centering
  \includegraphics[width=\columnwidth]{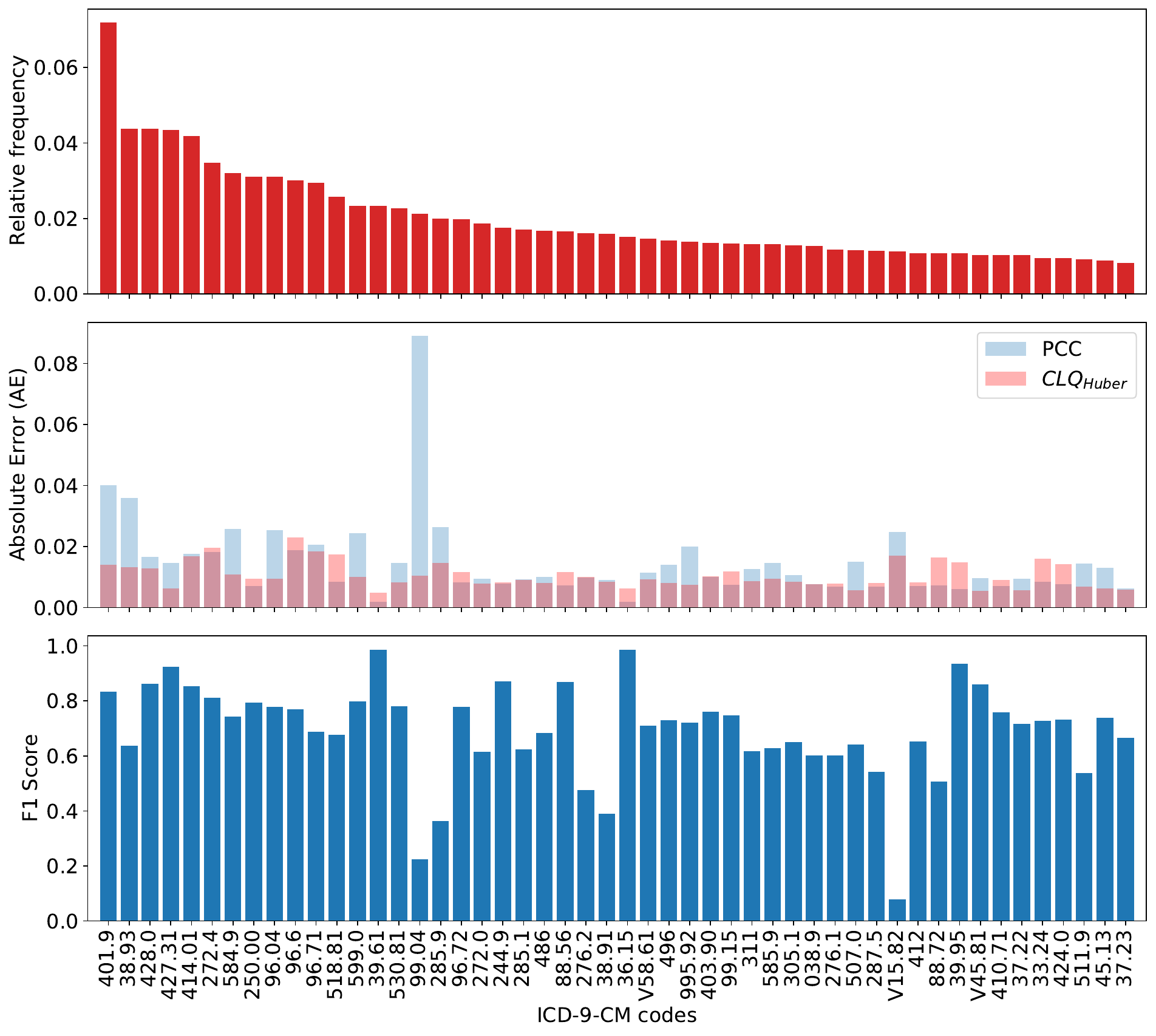}
  \caption{Relative frequency, absolute error, and F1 scores for each ICD code over MIMIC-III-50 dataset.}
    \label{figure:freqbarplot}
\end{figure}

Figure \ref{figure:freqbarplot} shows Absolute Error (AE) and F1 scores per ICD class, sorted by prevalence frequency over the MIMIC-III-50 dataset.

\begin{figure}[t!]
  \centering
  \includegraphics[width=\columnwidth]{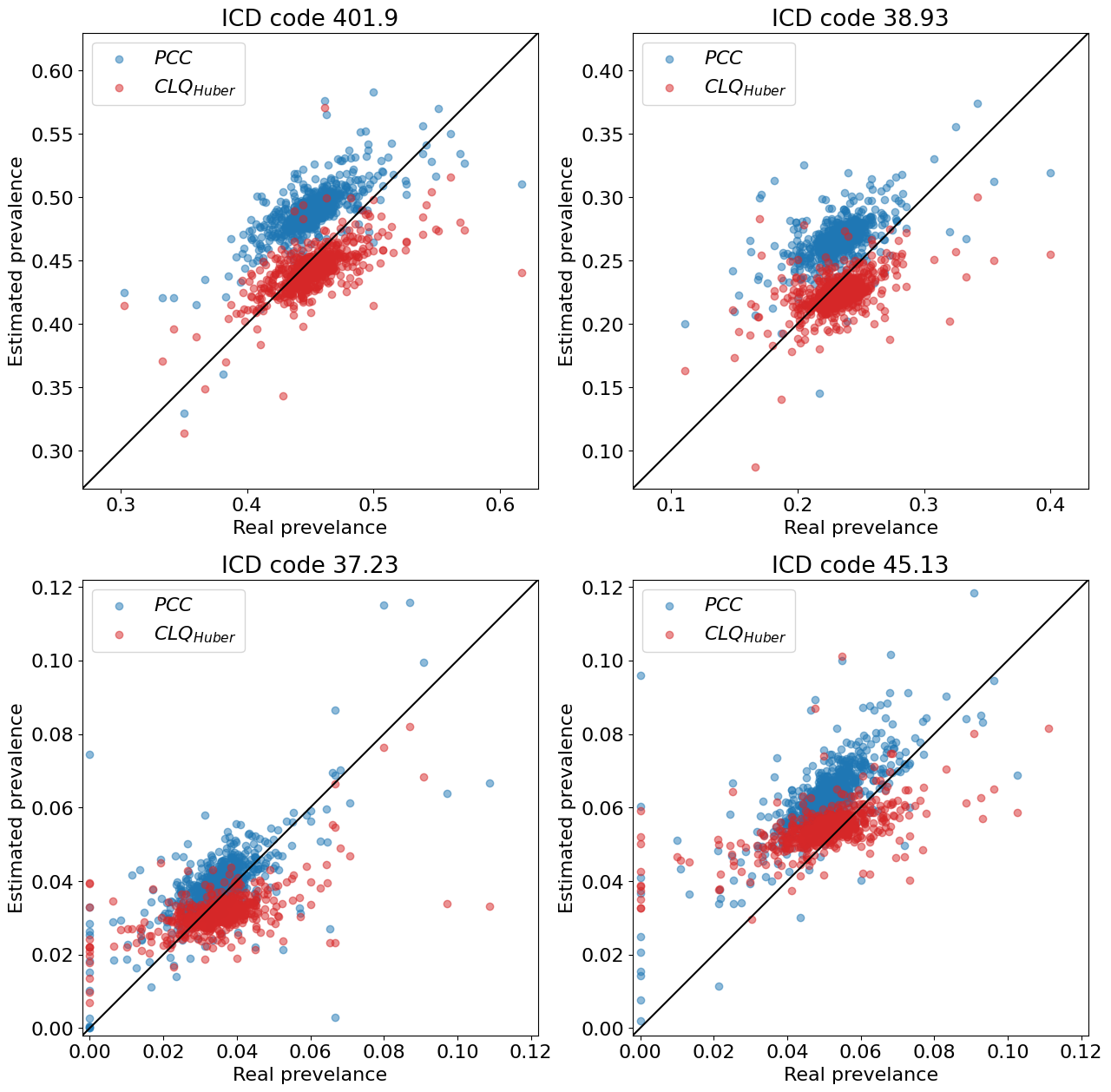}
  \caption{Estimated versus real prevalence for the two most frequent (top) and rarest (bottom) ICD codes in the MIMIC-III-50 dataset.}
    \label{figure:scatterplot}
\end{figure}

The results show that the $\text{CLQ}_\text{Huber}$ method outperforms PCC for nearly all ICD codes when it comes to accurately grasping ICD code prevalence. For instance, ICD code 401.9, which is the most frequent in the dataset, presents a high disparity in AE between PCC and $\text{CLQ}_\text{Huber}$ results. An even higher disparity is seen for code 99.04, which corresponds to {\it transfusion of packed cells} and has an average prevalence frequency. Further investigation revealed that despite having a high F1 score ($80.75$\%), the classification results for ICD code 401.9 have a notable difference between precision ($76.73$\%) and recall scores ($85.20$\%). This suggests that the classification model overestimates this class, perhaps due to its high frequency, resulting in inaccurate posterior probabilities. The $\text{CLQ}_\text{Huber}$ method appears to recognize this behavior and correct the results. Figure \ref{figure:scatterplot} aligns with the previous analysis, showing estimated versus real prevalence frequencies for frequent versus rare ICD codes (including code 401.9).

\end{document}